\documentclass[conf]{new-aiaa} 
\usepackage[utf8]{inputenc}
\usepackage{textcomp}

\usepackage{graphicx}
\usepackage{amsmath,bm}
\usepackage[version=4]{mhchem}
\usepackage{siunitx}
\usepackage{longtable,tabularx}
\usepackage{color}
\newcommand{\revv}[1]{ {\color{black}{#1}} }
\newcommand{\revb}[1]{ {\color{black}{#1}} }

\title{Stable Dynamic Mode Decomposition Algorithm for Noisy Pressure-Sensitive Paint Measurement Data}

\author{Yuya Ohmichi \footnote{Associate Senior Researcher, Aeronautical Technology Directorate}, Yosuke Sugioka \footnote{Researcher, Aeronautical Technology Directorate}, and Kazuyuki Nakakita \footnote{Senior Researcher, Aeronautical Technology Directorate}}
\affil{Japan Aerospace Exploration Agency, Tokyo 182-8522, Japan}

\begin{document}

\maketitle





\section{Introduction}
\lettrine{A}{N} increasing number of large-scale time-series datasets are being generated with the development of numerical and experimental techniques.
For example, pressure-sensitive paint (PSP) \cite{Steimle2012, Ali2016, Sugioka2018} and particle image velocimetry (PIV) \cite{Westerweel2013,Demauro2019,Singh2020} have been used for aerospace fluid analysis to obtain the spatial and temporal distribution of the flow field.
To gain deeper insights from multidimensional time-series data and utilize them for modeling fluid flow, data analysis techniques based on modal decomposition are being actively studied \cite{Holmes2012, Taira2017,Taira2020}.
Dynamic mode decomposition (DMD) \cite{Schmid2010} is one of the most commonly used modal analysis methods along with proper orthogonal decomposition (POD) \cite{Lumley1967}.
DMD has been applied in various studies \cite{Mariappan2014, Ali2016, Ohmichi2018, Ohmichi2019, Bai2020, Ranjan2020} due to its advantages in extracting both spatial modes and their associated temporal behavior.

In this study, we investigate the DMD method for noisy data, particularly for unsteady PSP measurement data.
PSP is an optical measurement method for the pressure field based on oxygen quenching of luminescence.
It has the advantage of enabling pressure measurements with high temporal and spatial resolution by using a fast-response porous binder.
However, when the pressure fluctuation of the observation target is small, such as in a flow field with low dynamic pressure, it is difficult to obtain a sufficiently large signal \cite{Liu2021}.
Therefore, when applying DMD to PSP data, a DMD algorithm is required that performs accurately and stably even in the presence of noise.

In DMD analysis, we define the input data matrix, which is a sequence of $m$ snapshots, as
$
    {\Psi} = [\psi_1, \psi_2,\cdots,\psi_{m}],
$
where $\psi_i$ $(i = 1,2,\cdots,m)$ is a column vector representing the $i$-th snapshot, and subscript $i$ corresponds to the time of each snapshot.
In addition, we define two matrices,
$
    X = [\psi_1, \psi_2,\cdots,\psi_{m-1}]
$
and
$
    Y = [\psi_2, \psi_3,\cdots,\psi_{m}].
$
In the standard DMD algorithm \cite{Schmid2010}, the DMD mode is defined as the eigenmode of the matrix $A = YX^+$, where $X^+$ is the pseudoinverse of $X$.
That is, $A$ is computed as the least-squares (LS) solution of $Y = AX$.

Various DMD methods \revv{utilizing noise-aware \cite{Dawson2016, Hemati2017, Taira2017, Nonomura2019}, sparse representation \cite{Jovanovic2014, Ohmichi2017c}, variable projection \cite{Askham2018}, and ODE-based \cite{Nonomura2021} approaches}, have been proposed.
In particular, Dawson et al. \cite{Dawson2016} \revv{and Hemati et al. \cite{Hemati2017}} pointed out that the LS regression used in standard DMD implicitly assumes that noise is only included in $Y$ and that there is no noise in $X$, which causes bias in the DMD eigenvalues if the input data matrix includes noise.
They proposed total least-squares (TLS) DMD that computes $A$ using TLS regression, which takes into account the noise in both $X$ and $Y$.
A number of studies \cite{Takeishi2017, Nonomura2019} also reported that TLS DMD outperformed standard DMD in terms of the accuracy of DMD eigenvalues.
However, it is well known that TLS regression is prone to computational instability in ill-conditioned problems.
Fierro et al. \cite{Fierro1997} mathematically derived the relationship between LS and TLS solutions using singular value decomposition (SVD) analysis and demonstrated that the contribution of components with small singular values can be larger in TLS solutions than in LS solutions for data with large noise.
In experimental measurement data, such as unsteady PSP data, the noise component may be too large to be ignored (i.e., the signal-to-noise ratio is small) when the signal component of the observation target is small.
In this case, the system to be solved by the DMD algorithm is likely to be ill-conditioned.
However, the computational stability of the TLS DMD algorithm has not been sufficiently investigated.

In this study, we investigate the stability of DMD algorithms to noisy data through DMD analysis of a numerical experiment and practical PSP measurement data.
We also \revv{apply the truncated TLS (T-TLS) regression and optimal truncation level selection proposed by Fierro et al. \cite{Fierro1997} to DMD algorithm.}
We evaluate the effectiveness of the \revv{T-TLS DMD} algorithm by comparing its results with those of other DMD algorithms.

\section{Algorithms for Dynamic Mode Decomposition}
\subsection{Standard and exact DMD} \label{standard_and_exact}
The typical algorithm for standard DMD is described below:
\begin{enumerate}
\setcounter{enumi}{0}
    \item Take the SVD of $X$, $X = U \Sigma V^T$.
    \item Compute the reduced order operator, $\tilde A = U^T Y V \Sigma^{-1}$.
    \item Solve the eigenvalue problem of $\tilde A$, $\tilde A \tilde \phi = \lambda \tilde \phi$.
    \item Every nonzero $\lambda$ is a DMD eigenvalue, and the corresponding DMD eigenvectors $\phi$ are given by $\phi = U \tilde \phi$.
\end{enumerate}

Tu et al. \cite{Tu2014} pointed out that standard DMD eigenvectors lie in the column space of $X$ but should lie in that of $Y$.
This can be achieved by computing $\phi$ as $\phi = \lambda^{-1} Y V \Sigma ^{-1} \tilde \phi$.
When this equation is used instead of Step 4 above, the algorithm is called exact DMD.
\revb{Note that the eigenvalues of the exact DMD are identical to those of the standard DMD. In this study, the exact DMD algorithm is used.}

\subsection{T-TLS DMD}
The algorithms for TLS DMD and T-TLS DMD are described below:
\begin{enumerate}
\setcounter{enumi}{0}
   \item Perform dimensionality reduction so that $r < m/2$ is satisfied, where $r$ is the reduced dimension.
   Letting $P_r$ be the first $r$ POD vectors of $\Psi$, the dimensionality reduction can be achieved by
    \begin{equation}
        \tilde X = P_r^T X, ~~\tilde Y = P_r^T Y. \label{lowdim}
    \end{equation}
   $P_r$ can be obtained by the SVD of $\Psi$.
   If $\Psi$ is too large to apply the batch SVD algorithm, online algorithms, such as incremental POD \cite{Ohmichi2017c}, can be applied.
   \item Construct the augmented data matrix $\tilde Z = \left[ \tilde X^T~\tilde Y^T\right]$, which is an $m-1$ by $2r$ matrix, and take the SVD of $\tilde Z$, $\tilde Z = U \Sigma V^T$.
   \item Partition the matrix $V$ such that
    \begin{equation}
        V = 
        \begin{bmatrix}
    	  V_{11} & V_{12} \\
    	  V_{21} & V_{22} 
    	\end{bmatrix}. \label{partV}
    \end{equation}
Here, $V_{11}$ and $V_{21}$ are $r$ by $k$ matrices, and $V_{12}$ and $V_{22}$ are $r$ by $q$ matrices (with $q = 2r - k$).
   \item Compute the reduced order operator, $\tilde A = V_{21} V_{11}^+$.
    \item Solve the eigenvalue problem of $\tilde A$, $\tilde A \tilde \phi = \lambda \tilde \phi$.
    \item Every nonzero $\lambda$ is a DMD eigenvalue and corresponding DMD eigenvectors $\phi$ are given by, $\phi = P_r \tilde \phi$.
\end{enumerate}
When the regularization parameter $k~(=1,2,\cdots,r)$ is $k=r$, the algorithm is identical to conventional TLS DMD.
In the above T-TLS DMD algorithm, the T-TLS regression proposed by Fierro et al. \cite{Fierro1997} is used instead of TLS regression.
By choosing the regularization parameter $k$ in (\ref{partV}) appropriately, the component corresponding to the small singular values that causes instability is truncated.
\revv{A similar truncation process has already been proposed by Hemati et al. \cite{Hemati2017} although they didn't introduce the parameter selection algorithm. They used the SVD of the augmented data matrix $Z = \left[  X^T~ Y^T\right]$ as a pre-processing (dimensionality reduction), while in the proposed algorithm, the SVD of the data matrix $\Psi$ is used as a pre-processing and the truncation including optimal parameter selection is performed in the low-dimensional space.}

The choice of the optimal $k$ is nontrivial; in this study, we use $k$ that minimizes the error $E(k) = \|\tilde Y -\tilde A\tilde X\|_2$ as a simple criterion \cite{Fierro1997}.
That is, we calculate $E(k)$ for all $k$ and adopt the $k$ corresponding to the smallest $E(k)$ as the optimal parameter.
\revv{In the presence of noise, $E(k)$ does not necessarily decrease monotonically as $k$ increases because TLS regression does not necessarily minimize the LS error.}

\section{Results and Discussion}
\subsection{Numerical test}
\begin{figure}[hbt!]
\centering
\includegraphics[width=.95\textwidth]{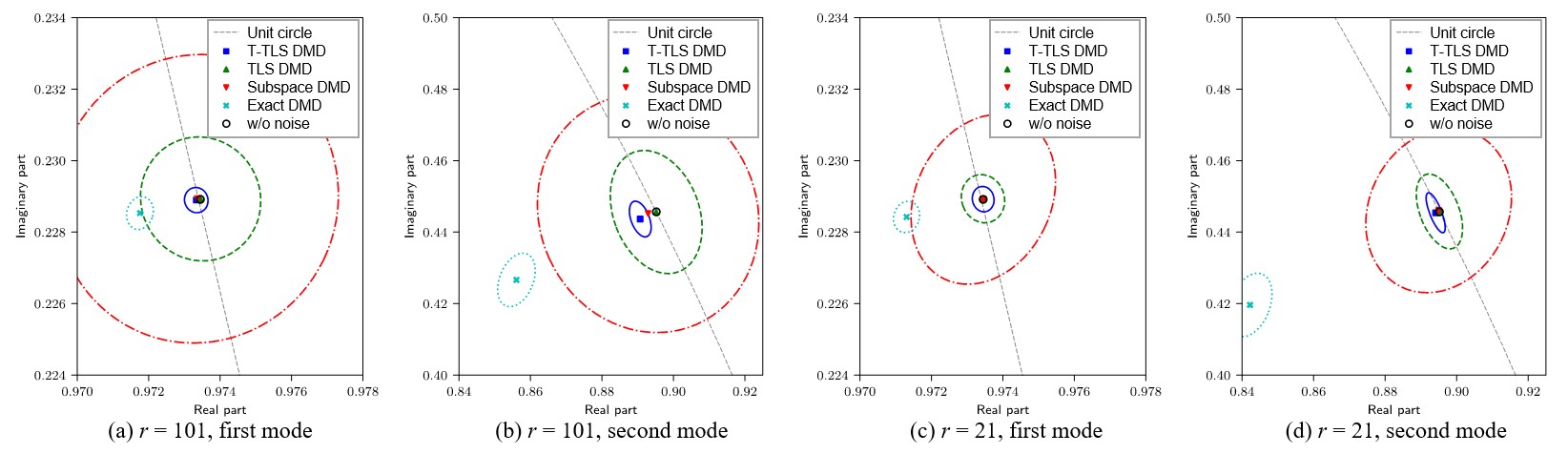}
\caption{Comparison of eigenvalues estimated by T-TLS, TLS, subspace, and \revb{exact} DMD algorithms \revv{with $\sigma^2 = 0.1$}.
}\label{cylevals}
\end{figure}

\begin{figure}[hbt!]
\centering
\includegraphics[width=.85\textwidth]{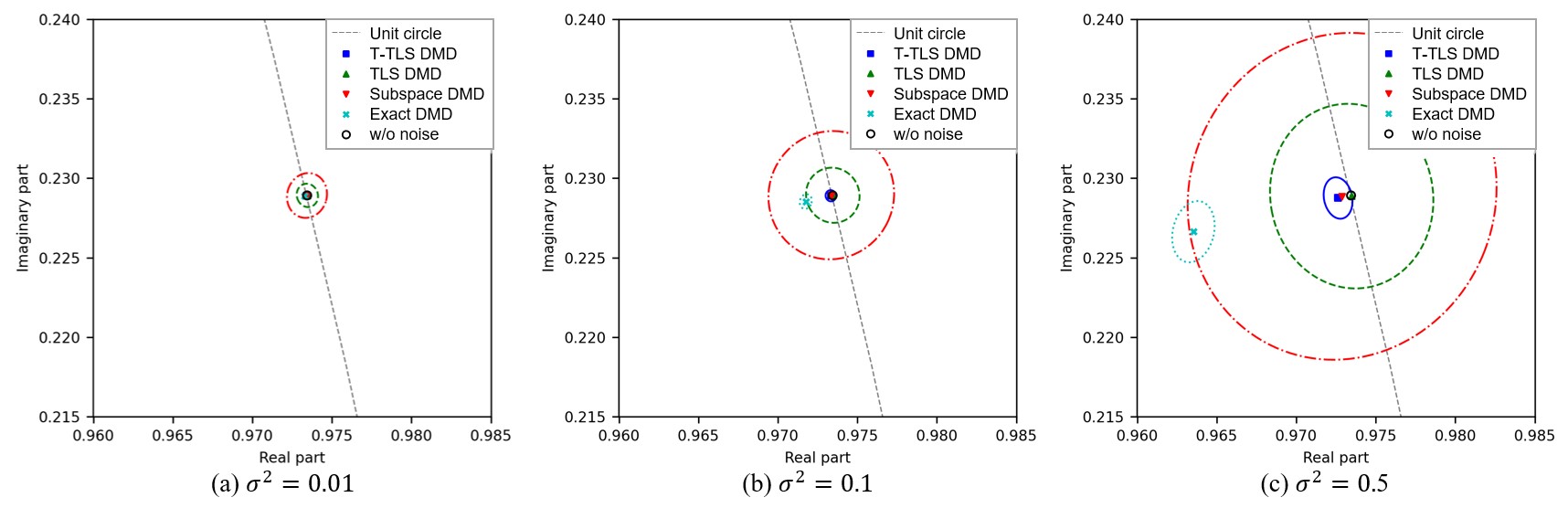}
\caption{Effects of noise variance on eigenvalues of the K\'{a}rm\'{a}n vortex mode with $r=101$, estimated by T-TLS, TLS, subspace, and \revb{exact} DMD algorithms. 
}\label{effect_of_noise}
\end{figure}

\begin{figure}[hbt!]
\centering
\includegraphics[width=.95\textwidth]{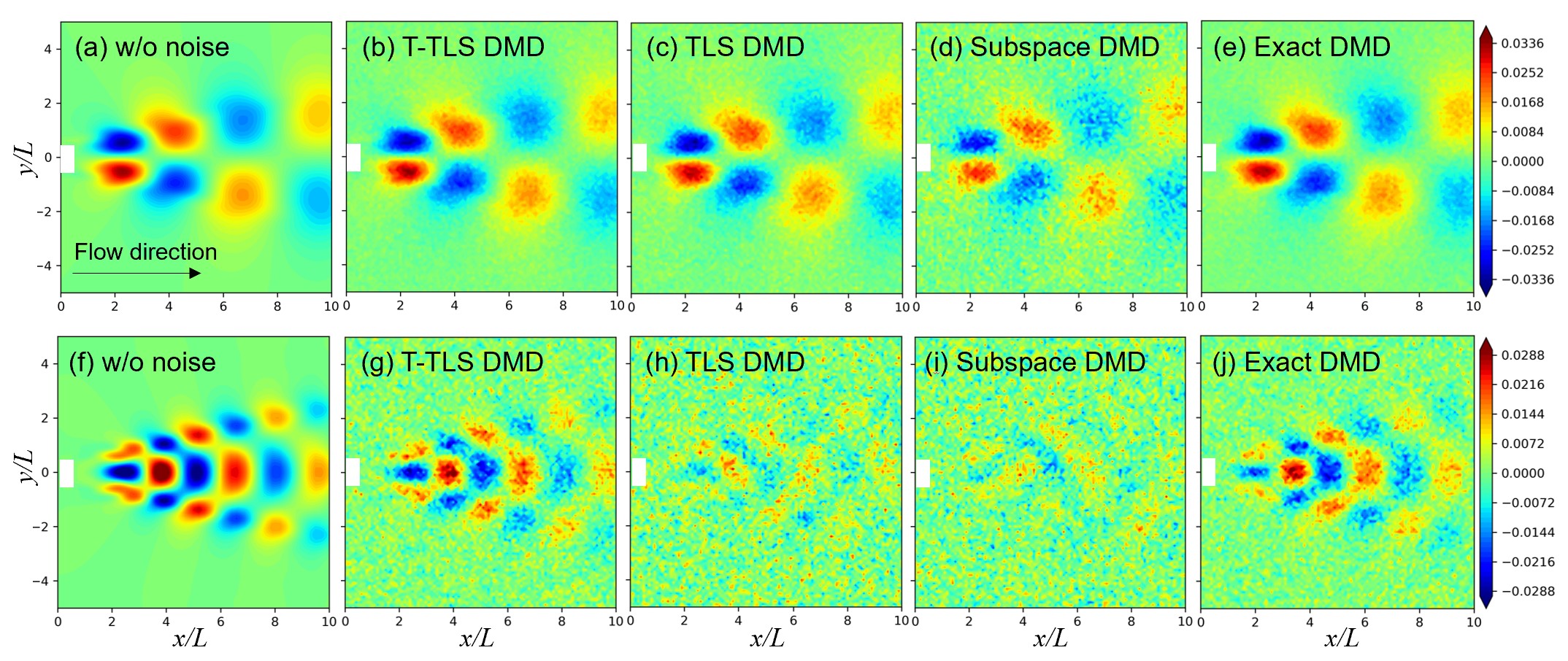}
\caption{Eigenvectors for the K\'{a}rm\'{a}n vortex mode (upper row) and second harmonic mode (lower row) estimated by T-TLS, TLS, subspace, and \revb{exact} DMD algorithms \revv{with $\sigma^2 = 0.1$}. 
\revv{The eigenvectors estimated by standard DMD without noise are also shown as a reference.}
} \label{cylevec}
\end{figure}
First, the performance of the proposed and existing methods in the DMD analysis of the K\'{a}rm\'{a}n vortex shedding phenomenon behind a square cylinder was investigated.
The unsteady flow field obtained by a two-dimensional numerical fluid simulation was used as input data.
The Mach number was 0.2 and the Reynolds number was 100.
The computational grid was a Cartesian grid of $511 \times 381$ points (in the uniform flow and transverse directions),
and the cylinder had its center located at the origin $(x/L, y/L) = (0, 0)$.
Here, $L$ is the length of the side of the square cylinder. 
The sixth-order accurate compact finite difference method \cite{Lele1992,Visbal2002} and eighth-order accurate filtering \cite{Gaitonde2000} were used for spatial discretization, and the third-order three-step TVD Runge--Kutta method \cite{Shu1988} was used for time integration.
As input datasets to DMD, the velocity field in the wake region $x/L = [0,~10]$ and $y/L = [-5,~5]$ were sampled into equally spaced $101 \times 101$ grid points.
The number of snapshots was 400, and the time interval between the snapshots was $\Delta t = 0.25 L/U_\infty$.
To investigate the effect of noise, random normal noise with variance $\sigma^2$ was added to each snapshot as observation noise.
The input data (i.e., velocity) and noise amplitudes were normalized by the uniform flow velocity, $U_\infty$.
The input data were projected onto the $r$ POD vectors using (\ref{lowdim}).

Figure \ref{cylevals} presents the eigenvalues, \revv{$\lambda$,} of the first  K\'{a}rm\'{a}n vortex ($St = 0.147$) and second ($St = 0.296$) modes obtained by the T-TLS, TLS, subspace, and \revb{exact} DMD algorithms.
\revb{As mentioned in Sec. \ref{standard_and_exact}, the eigenvalues of the exact DMD are identical to those of the standard DMD.}
The subspace DMD \cite{Takeishi2017} is one of the state-of-the-art DMD algorithms. 
The Strouhal number, $St$, represents the frequency nondimensionalized by $L$ and $U_\infty$.
The figure displays the 95\% confidence ellipse and the average eigenvalue calculated based on 1,000 random trials with observation noise of $\sigma^2 = 0.1$.
The 95\% confidence ellipse indicates the magnitude of the variation of the computed eigenvalues, and the smaller the confidence ellipse, the more stable the algorithm.
The eigenvalues of the noiseless case (obtained by standard DMD) are also plotted as true values.
The optimal regularization parameters for T-TLS DMD were approximately $k=45$ and 5 for $r=101$ and 21, respectively.
It was observed that for the case of the reduced dimension $r = 101$, the eigenvalues of \revb{exact} DMD had a larger deviation from the true eigenvalues than the other methods.
The eigenvalues were shifted inward from the circumference of the unit circle, indicating that the growth rate decreased due to noise.
In contrast, the average eigenvalues of T-TLS, TLS, and subspace DMD had small deviations from the true values; in particular, the average eigenvalue of TLS DMD was almost identical to the true value.
However, the 95\% confidence ellipse for TLS and subspace DMD was large, indicating that the variation of the eigenvalues caused by noise was large.
In the T-TLS DMD, the variation of the eigenvalues was superior to that of TLS DMD, and it was confirmed that the T-TLS DMD algorithm was stable due to the effect of truncation.
The 95\% confidence ellipse for \revb{exact} DMD was also relatively small.
The eigenvalue of T-TLS DMD for the second mode (Fig. \ref{cylevals}b) indicates that the eigenvalue was slightly shifted to the damping side.
This was due to truncation in the T-TLS algorithm, which attenuated part of the signal representing the second mode.
Figures \ref{cylevals}c and d display the results for the reduced dimension $r=21$.
Although the variation of the eigenvalues obtained by each DMD method was reduced by a decrease in the reduced dimension, the T-TLS DMD still had the smallest variation.
\revv{
Figure \ref{effect_of_noise} presents the estimated eigenvalues at several noise levels.
The larger the noise, the larger the variation in the eigenvalues, but the T-TLS DMD was the most stable at all noise levels.
The eigenvalues of T-TLS and subspace DMD were found to be slightly shifted to the damping side for the case of the largest noise level $\sigma^2=0.5$.
}

Figure \ref{cylevec} presents a comparison of the DMD eigenvectors obtained by each algorithm.
To compare the robustness of each DMD algorithm, the eigenvector with the largest error among the results of random trials with noise is displayed.
\revv{For ease of comparison, the phases of all modes have been adjusted to match by multiplying by a complex number \cite{Nonomura2018}.}
For the K\'{a}rm\'{a}n vortex mode, all methods exhibited similar distributions; however, the subspace DMD result was somewhat unclear.
For the second mode, the results for TLS and subspace DMD had unclear distributions.
That is, the coherent structures behind the cylinder was collapsed.
In contrast, the eigenvector of T-TLS DMD was clear, indicating that by the regularization effect of truncation, the T-TLS DMD algorithm was stable even when the TLS DMD algorithm was unstable.
The eigenvectors of the \revb{exact} DMD were calculated stably, although there was bias in the eigenvalues.
From this analysis, we found that T-TLS DMD calculated DMD eigenvalues and eigenvectors more stably and accurately than other methods even in the presence of observation noise.

\subsection{Pressure-sensitive paint measurement data}
\begin{figure}[hbt!]
\centering
\includegraphics[width=.3\textwidth]{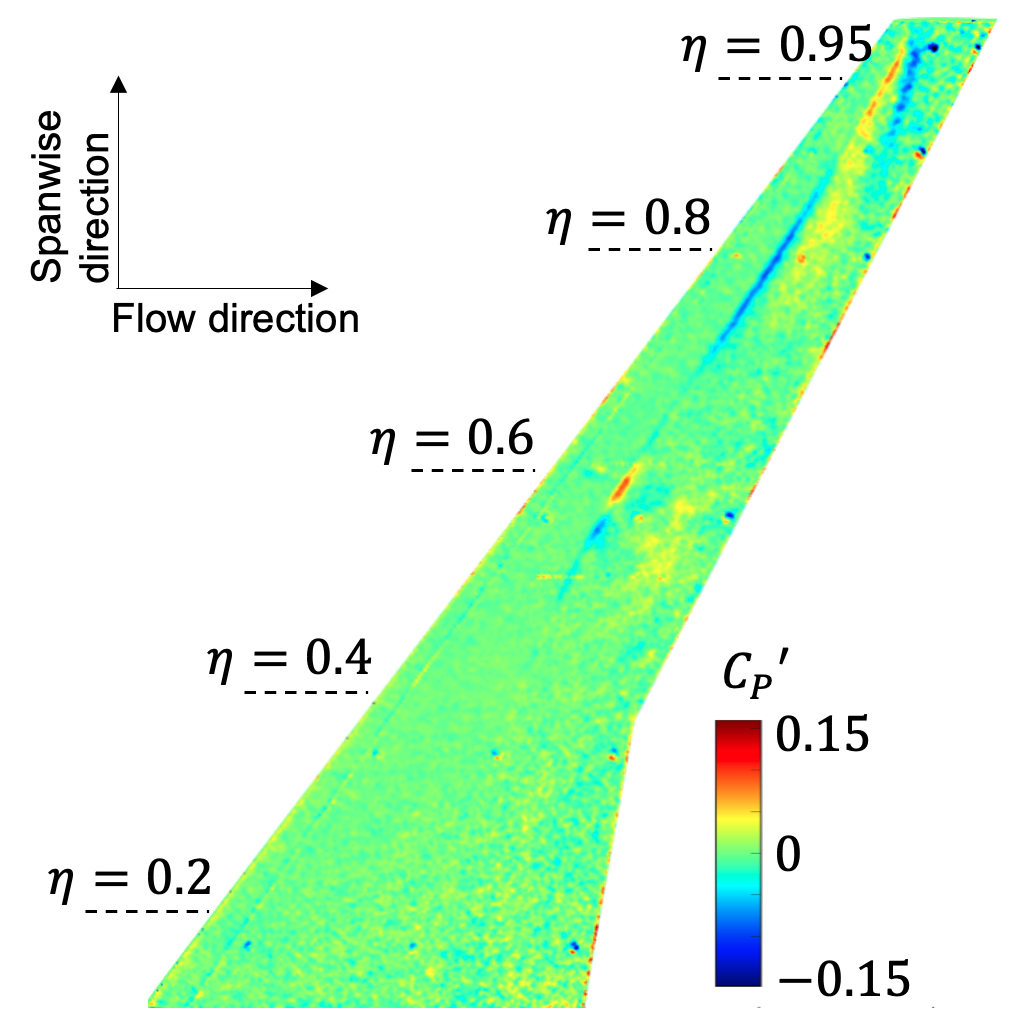}
\caption{Pressure fluctuation ($C_P'$) distribution of the instantaneous flowfield obtained by pressure-sensitive paint measurement. } \label{PSPins}
\end{figure}

Next, we tested the effectiveness of the T-TLS DMD on practical PSP measurement data.
We analyzed the transonic buffet phenomenon appearing on the NASA CRM wing surface \cite{Sugioka2021, Koike2016}.
The experiment was conducted in a 2m $\times$ 2m transonic wind tunnel (JTWT1) at the JAXA Chofu Aerospace Center.
The Mach number was set to 0.85, and the Reynolds number based on the mean aerodynamic chord was set to $2.27 \times 10^6$.
The angle of attack was set to $3.71^\circ$.
The unsteady pressure distribution over the wing surface was measured using polymer/ceramic PSP (PC-PSP) developed by Sugioka et al. \cite{Sugioka2018}.
\revv{
Evaluating the noise level based on the previous study \cite{Sugioka2021}, the signal-to-noise ratio was approximately 20 dB for large amplitude fluctuations at the foot of the shockwave, and $ O(1)$ dB for fluctuations downstream of the shockwave.
}
For details of the experiment, please refer to \cite{Sugioka2021}.

Figure \ref{PSPins} presents the instantaneous distribution of pressure fluctuation $C_P'$ obtained from the experiment.
$C_P'$ represents the instantaneous pressure coefficient after subtracting the time-averaged value.
Sugioka et al. \cite{Sugioka2021} demonstrated that in this flow, the shockwave generated on the wing oscillated and a pressure fluctuation pattern, the so-called buffet cell, appeared.
The pressure pattern occurring near the spanwise position $\eta = 0.6$ in Fig. \ref{PSPins} is a buffet cell.
In addition, pressure patterns that are not buffet cells can also be seen in this figure.
In particular, the mottled distribution over the entire wing and the relatively large fluctuation at the model edges are pseudo pressure distributions (i.e., noise).
The former was caused randomly by camera shot noise and was pronounced in regions with small pressure fluctuations.
The latter was due to alignment errors in the PSP process, which were caused by vibrations of the model.

The pressure coefficient distribution obtained in this experiment was input to the DMD analysis.
The number of snapshots was 500, and the time interval between snapshots was $\Delta t = 0.38 c_{\rm MAC}/U_\infty$.
The reduced dimension was set to $r=101$.
The optimal regularization parameter for T-TLS DMD was $k=49$.
\revv{The difference in computation time between DMD algorithms was about 1\% of the total computation time. This is because in DMD analysis of high-dimensional data such as spatial distribution data, most of the computational cost is consumed by the SVD-based subspace projection, which is a process common to all DMD algorithms.}

\begin{figure}[hbt!]
\centering
\includegraphics[width=.3\textwidth]{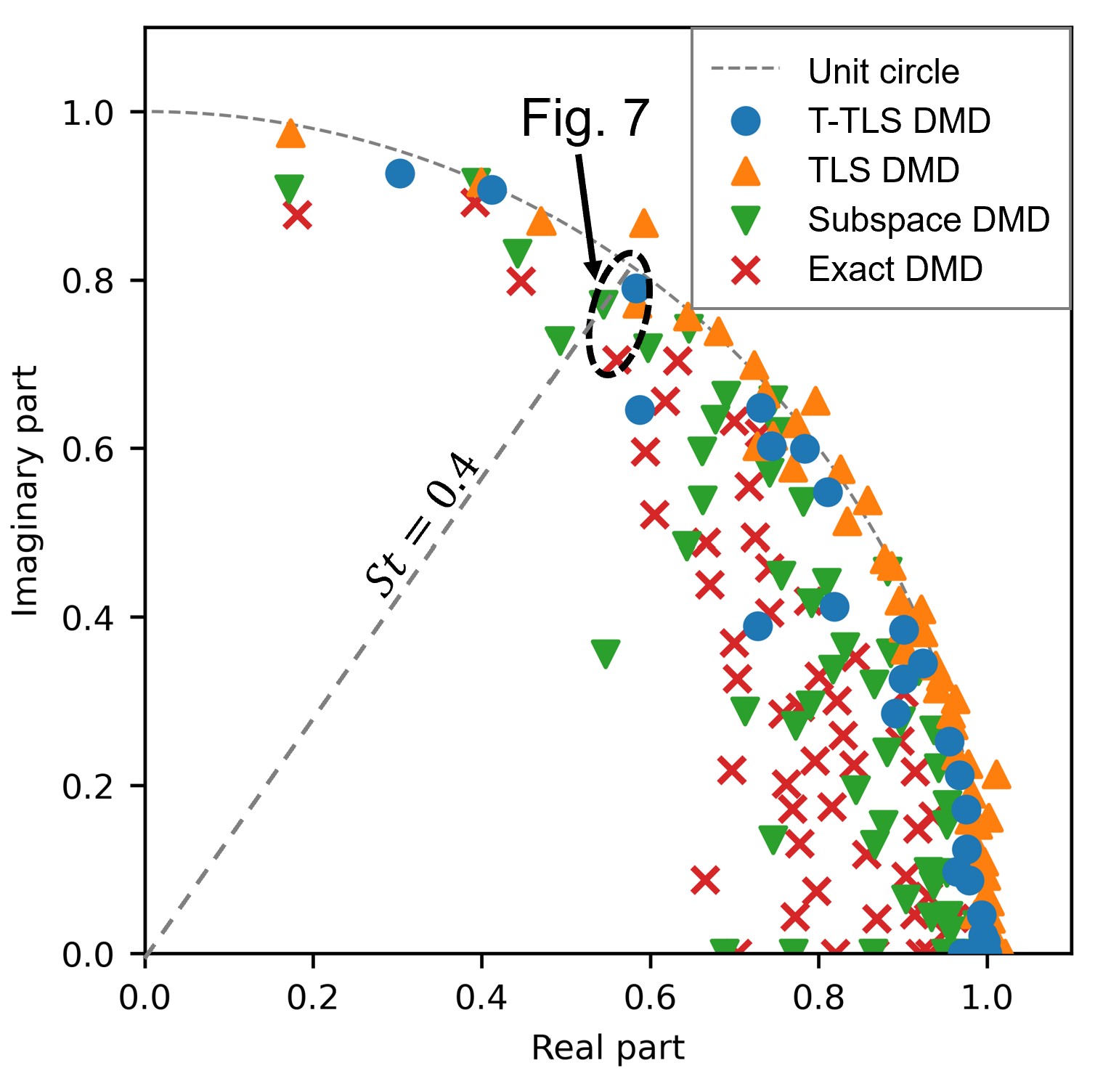}
\caption{Eigenvalue distributions obtained by T-TLS, TLS, subspace, and exact DMD algorithms.} \label{PSPevals}
\end{figure}

Figure \ref{PSPevals} presents the eigenvalues, \revv{$\lambda$,}  obtained using the T-TLS, TLS, subspace, and \revb{exact} DMD algorithms.
It can be seen that the distribution of eigenvalues differed greatly depending on the DMD algorithm used.
Comparing the results of T-TLS and TLS DMD, the eigenvalues of TLS DMD had absolute values greater than 1, indicating that the growing mode appeared, whereas the eigenvalues of T-TLS DMD did not exhibit the growing mode.
Most of the eigenvalues of T-TLS DMD were distributed near the unit circle; that is, they were expressed as modes close to steady oscillation.
In contrast, the eigenvalues of subspace and \revb{exact} DMD were distributed inside the unit circle, and many of the modes were damping modes.
\revv{Figure \ref{changek} presents a dependence of T-TLS DMD eigenvalues on the regularization parameter $k$.
When $k = 101$, the algorithm is identical to TLS DMD algorithm.
As $k$ decreased, some eigenvalues appeared on the damping side, while most of the eigenvalues were still distributed near the unit circle.}

\begin{figure}[hbt!]
\centering
\includegraphics[width=0.9\textwidth]{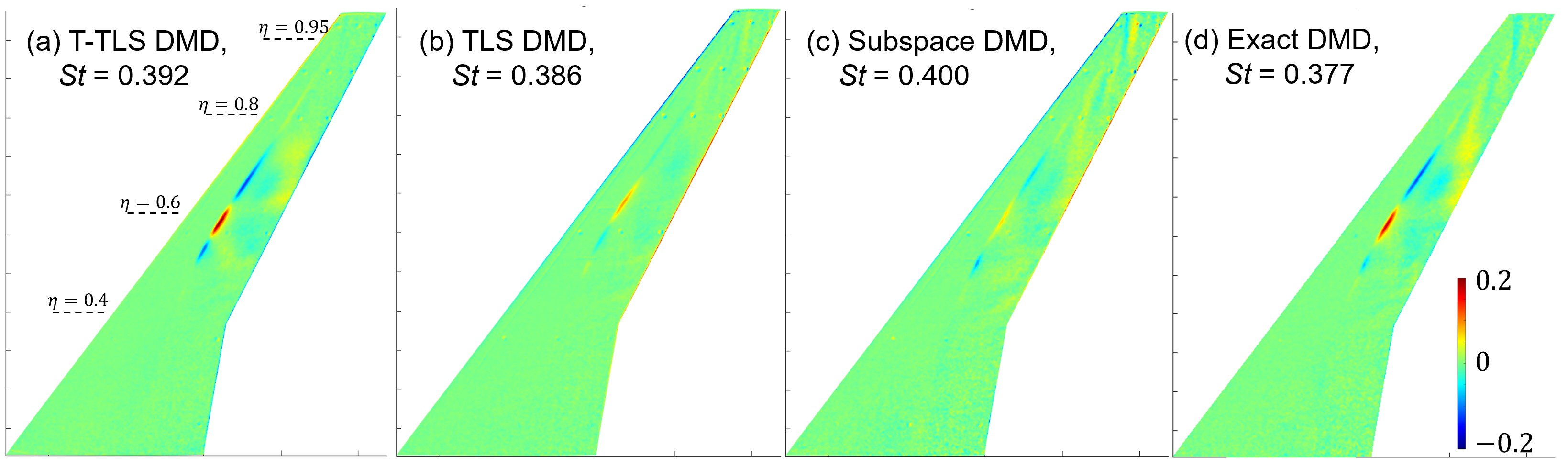}
\caption{\revv{Real parts of} eigenvectors estimated by T-TLS, TLS, subspace, and \revb{exact} DMD algorithms. Modes with the frequency of $St \approx 0.4$ are displayed. 
} \label{PSPevecs}
\end{figure}

Figure \ref{PSPevecs} presents a comparison of the DMD eigenvectors corresponding to $St \approx 0.4$ \revv{indicated in Fig. \ref{PSPevals}}.
According to Sugioka et al. \cite{Sugioka2021} and Ohmichi et al. \cite{Ohmichi2018}, $St \approx 0.4$ is included in the range of the characteristic frequency of buffet cells.
Since the eigenvalue distribution differed depending on the algorithm, as illustrated in Fig. \ref{PSPevals}, the modes close to $St = 0.4$ for each DMD algorithm are displayed. 
Figure \ref{PSPevecs}a demonstrates that a clear buffet cell pattern was captured by T-TLS DMD \revv{($k = 49$)}.
That is, a pressure fluctuation pattern with periodicity in the spanwise direction appeared near $\eta \approx 0.6$.
\revv{This pressure fluctuation pattern has also been observed in numerical fluid simulations \cite{Ohmichi2018}.
The eigenvectors ($St\approx 0.4$) of T-TLS DMD with $k=25$ and 75 were similar to Fig. \ref{PSPevecs}a while a slightly larger noise appeared for $k=75$ (not shown here).}
Exact DMD also extracted a clear buffet cell pattern.
Exact DMD was thus considered to be a relatively stable algorithm although it had the problem of eigenvalues shifting to the damping side.
Although a similar buffet cell pattern appeared in the eigenvector of TLS and subspace DMD, it was more obscure than that of T-TLS DMD.
This may be due to the overfitting of the TLS and subspace DMD algorithms to the noise.
\revv{TLS and subspace DMD had other $St \approx 0.4$ modes not shown in Fig. \ref{PSPevecs}, but their distributions also contained noise.}

\revv{
\begin{figure}[hbt!]
\centering
\includegraphics[width=.3\textwidth]{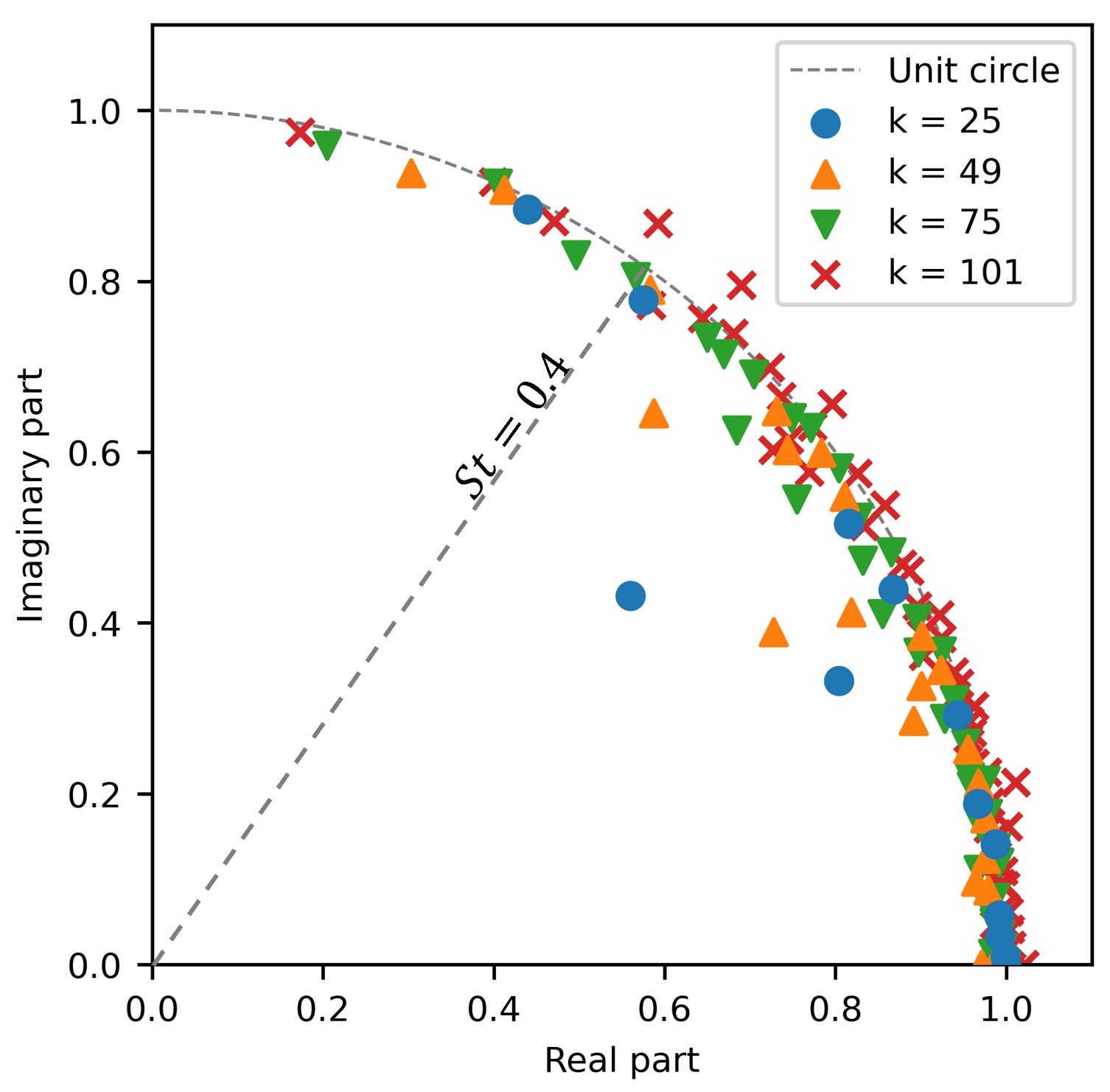}
\caption{Eigenvalue distributions obtained by T-TLS DMD with changing the regularization parameter $k$.} \label{changek}
\end{figure}
}

\section{Conclusion}
In this study,
\revv{we investigated the stability of DMD algorithms to noisy data.
To achieve a stable DMD algorithm, we applied the truncated TLS (T-TLS) regression and optimal truncation level selection to the TLS DMD algorithm.}
By adding truncation regularization to the TLS DMD algorithm, T-TLS DMD improves the stability of the computation while maintaining the accuracy of TLS DMD.
The effectiveness of the T-TLS DMD was evaluated by the analysis of the wake behind a cylinder and PSP data for the buffet cell phenomenon.
The results showed the importance of regularization in the DMD algorithm.
With respect to the eigenvalues, T-TLS DMD was less affected by noise, and accurate eigenvalues could be obtained stably, whereas the eigenvalues of TLS and subspace DMD varied greatly due to noise.
It was also observed that the eigenvalues of the standard and exact DMD had the problem of shifting to the damping side, as reported in previous studies.
With respect to eigenvectors, T-TLS and exact DMD captured the characteristic flow patterns clearly even in the presence of noise, whereas TLS and subspace DMD were not able to capture them clearly due to noise.



\section*{Acknowledgments}
This work was supported in part by the Japan Society for the Promotion of Science (JSPS) KAKENHI (grant no. 20K14958). 

\bibliography{sample}

\end{document}